# K-Means Segmentation of Alzheimer's Disease In Pet Scan Datasets – An Implementation

Meena A[1], Raja K[2]

Research Scholar, Sathyabama University, Chennai, India
Principal, Narasu's Sarathy Institute of Technology, Salem, India
Kabimeena2@hotmail.com, raja_koth@yahoo.co.in

**Abstract.** The Positron Emission Tomography (PET) scan image requires expertise in the segmentation where clustering algorithm plays an important role in the automation process. The algorithm optimization is concluded based on the performance, quality and number of clusters extracted. This paper is proposed to study the commonly used K- Means clustering algorithm and to discuss a brief list of toolboxes for reproducing and extending works presented in medical image analysis. This work is compiled using AForge .NET framework in windows environment and MATrix LABoratory (MATLAB 7.0.1)

## 1 Introduction

Positron Emission Tomography (PET) detects chemical and physiological changes related to functional metabolism [1, 7]. This scan images are more sensitive than other image techniques such CT and MRI because the other imaging techniques only shows the physiology of the body parts.

This paper presents a study on the application of well known K-Means clustering algorithm. This algorithm is used to automate process of segmentation of the tumor affected area based on the datasets classified by its type, size, and number of clusters [2]. The rest of the paper is organized as follows. Section 2 states the related work in this area; section 3 describes the K-Means clustering algorithms, section 4 and section 5 presents the implementation methods. Comparison between AForge .NET framework and MATLAB are made in the concluding section.

## 2 Related Work

Digital image processing allows an algorithm to avoid problems such as the build-up of noise and signal distortion occuring in analog image processing.

Fulham *et al*. (2002) stated that quantitative positron emission tomography provides the measurements of dynamic physiological and biochemical processes in humans. Ciccarelli *et al.* (2003) and Meder *et al.* (2006) proposed a method to identify sclerosis that disrupts the normal organization or integrity of cerebral white matter and the underlying changes in cartilage structure during osteoarthritis. Functional imaging methods are also being used to evaluate the appropriateness and efficacy of therapies such as Parkinson's disease, depression, schizophrenia, and Alzheimer's disease [5]. Quantum dots (qdots) are fluorescent nano particles of semiconductor material which are specially designed to detect the biochemical markers of cancer (Carts-Powell, 2006). Osama (2008) explained about the comparision of clustering algorithms and its application based on the type of dataset used. In 2009, Stefan *et al.* described the structured patient data for the analysis of the implementation of a clustering algorithm. These authors correlated images of the dementia affected brain with other variables, for instance, demographic information or outcomes of clinical tests [6]. In this paper, clustering is applied to whole PET scans.

## 3 K-Means Clustering Algorithm

Clustering is used to classify items into identical groups in the process of data mining. It also exploits segmentation which is used for quick bird view for any kind of problem. K-Means is a well known partitioning method. Objects are classified as belonging to one of k groups, k chosen a priori [3]. Cluster membership is determined by calculating the centroid for each group and assigning each object to the group with the closest centroid. This approach minimizes the overall within-cluster dispersion by iterative reallocation of cluster members [4].

Pseudo code for centroid calculation

```
Step1: Initialize / Calculate new centroid
Step2: Calculate the distance between object and every
       centroid
Step3: Object Clustering
Step4: If any object moved from one cluster to the other,
       go to step1 or Stop
```

Pseudo code for image segmentation

```
Step1: Initialize centroids corresponding to required
       number  of clusters
Step2: Calculate original centroid  (Call K- Means)
Step3: Calculate the mask
Step4: Do the segmentation process
```

## 4 AForge .NET Framework

### 4.1 Text Conversion

The sample data was collected from Alzheimer's Disease Neuroimaging Initiative (ADNI) [9, 10]. The given input image consisting of various pixels points is then converted to byte stream. The converted byte stream is stored in a jagged array in the form of a text file as datasets.

### 4.2 Data Preprocessing and Image Retrieval

The dataset in the form of text file is imported into the system then it is converted in the Comma Separated Values (CSV) format. The converted CSV in byte stream is used to initialize the jagged pixel array and the array is transformed to bitmap image. Using jagged array the length of each array pixel can be adjusted. It can use less memory and be faster than two dimensional arrays because of uneven shape.

The image retrieved from the byte stream is shown in the form of the thumbnail view in this part of the system and the system's mode displays the rows and columns. This jagged index is used to reallocating the pixel values to its corresponding centroid values.

### 4.3 Knowledge based cluster analysis

This part of the system is to select the K- Means clustering algorithm for the given dataset and to segment an image. Randomly the number of cluster is selected as 5. The clustering algorithm is used to automate the process of segmentation here when the clustering is done based on the pixel values. It can be changed to its pixel values so that the image segmentation is made possible in the byte stream.

## 5 Matlab

The given image image is loaded in to MATLAB. First the number of clusters is assigned. Then the centroid (c) initialization is calculated as follows

$$c = (1:k)*m / (k+1) \qquad (1)$$

where the double precision image pixel in single column (m) value and number of centroid (k) is used to calculate the initial centroid value

The calculation of distance (d) between centroid and object is derived from

$$d = abs\,(o\,(i) - c) \qquad (2)$$

Equation (2) o(i) is known as one dimensional array distance. Using that value, new centroid is calculated in equation 3

$$nc(i) = sum\ (a.*h(a)) / sum\ (h(a)) \qquad (3)$$

where the value object clustering function (a) and non zero element obtained from object clustering h(a) is used to compute the new centroid (nc) value. The resultant new centroid value is used for masking creation and then the image segmentation.

## 6  Results on synthetic image

Figs. 1 are the segmenting results on a synthetic image using K-Means clustering algorithms in AForge .NET and MATLAB.

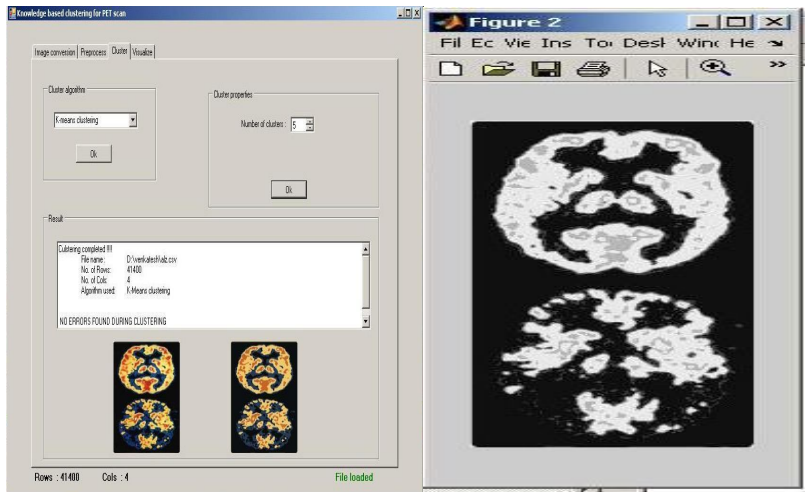

(a)　　　　　　　　　　　　　　(b)

**Fig. 1.** Segmentation results on synthetic image. (a) AForge .NET result. (b) MATLAB result.

The obtained image from MATLAB and .NET framework is analyzed using Medical Image Processing And Visualization (MIPAV) tool. The selection volume of interest is identified and the statistical parameter is listed below. Fig. 2 shows that the co efficient of variance value in .NET framework is lesser than MATLAB environment and it is the proof for less significant distribution in .NET

| S. No | Parameter | Result obtained from | |
|---|---|---|---|
| | | MATLAB | .NET |
| 1. | Average | 86.0916 | 79.2168 |
| 2. | Standard Deviation | 92.0758 | 65.3007 |
| 3. | Coefficient of variance | 106.951 | 82.433 |

(a)

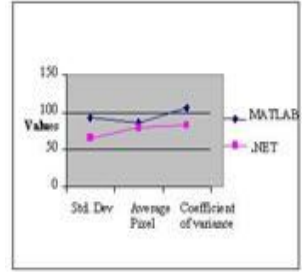

(b)

**Fig. 2.** Results obtained from Aforge .NET and MATLAB. (a) Statistical Values. (b) Graph representation

## 7 Conclusion and Future work

This paper deals with the basic K-Means algorithm in different working platform. First the K-Means is tested on AForge.NET framework in windows environment. Then the obtained image is compared with the image obtained in MATLAB environment. AForge .NET environment produced the optimal segmented image when compared to that got from MATLAB. In future the real PET image datasets with exact CPU utilization is to be studied.

## References

1. Koon-Pong Wong., Dagan Geng., Steven R. Meikle., Michael J.Fulham.: Segmentation of Dynamic PET Images Using cluster analysis. *IEEE Transactions on nuclear science, Vol. 49, 200-207( 2002)*
2. Andreas Hapfelmeier., Jana Schmidt., Marianne Muller., Stefan Kramer.: Interpreting PET scans by structured Patient Data: A Data mining case study in dementia Research. *IEEE Knowledge and Information Systems,213-222( 2009)*
3. M.C. Su., C. H. Chou.: A Modified Version of the K – Means Algorithm with a Distance Based on Cluster Symmetry. IEEE Trans. On Pattern Analysis and Machine Intelligence, *vol*.23, 674 – 680( 2001)
4. Oyelade., O. J, Oladipupo., O. O, Obagbuwa., I. C.: Application of k-Means Clustering algorithm for prediction of Students' Academic Performance, *International Journal of Computer Science and Information Security, vol. 7,292-295( 2010)*
5. Brookmeyer., R; Johnson., E; Ziegler-Graham., K; Arrighi, HM.: Forecasting the global burden of Alzheimer's disease, 186 – 91(2007)
6. D. L. Pham., C. Xu., L. Prince.: Current methods in medical images segmentation, Annual review of biomedical engineering, *vol*.2, 315-337( 2000)
7. H. N. Wagber., Z. Szabo., J. W. Buchanan.: Principles of nuclear medicine, Pensylvania, 564 – 575(1995)